\documentclass[twoside]{article}

\usepackage[accepted]{aistats2021}
\usepackage[round]{natbib}

\usepackage{hyperref}
\usepackage{url}
\usepackage{algorithm}
\usepackage{graphicx}
\usepackage{natbib}
\usepackage[noend]{algpseudocode}
\usepackage{listing}
\usepackage{chngpage}
\usepackage{subcaption}
\usepackage[T1]{fontenc}
\usepackage{multirow}
\usepackage{booktabs}
\usepackage[switch]{lineno}
\usepackage{microtype}
\usepackage{amssymb}
\usepackage{amsmath,amsfonts}
\usepackage{amsopn,bm,mathtools}

\usepackage{nicefrac}       
\newcommand{\vct}[1]{\boldsymbol{#1}} 
\newcommand{\mat}[1]{\boldsymbol{#1}} 

\newcommand{\field}[1]{\mathbb{#1}}
\newcommand{\R}{\field{R}} 
\newcommand{\T}{^{\textrm T}} 

\newcommand{\ProbOpr}[1]{\mathbb{#1}}

\newcommand{\expect}[2]{%
\ifthenelse{\equal{#2}{}}{\ProbOpr{E}_{#1}}
{\ifthenelse{\equal{#1}{}}{\ProbOpr{E}\left[#2\right]}{\ProbOpr{E}_{#1}\left[#2\right]}}} 
\newcommand{\var}[2]{%
\ifthenelse{\equal{#2}{}}{\ProbOpr{VAR}_{#1}}
{\ifthenelse{\equal{#1}{}}{\ProbOpr{VAR}\left[#2\right]}{\ProbOpr{VAR}_{#1}\left[#2\right]}}} 

\DeclareMathOperator{\argmin}{arg\,min}

\newcommand{\parde}[2]{\frac{\partial #1}{\partial  #2}}

\newcommand{\vx}{{\vct{x}}}

\newcommand{\mI}{\mat{I}}

\newcommand{\sA}{\mathcal{A}}

\newcommand{\sL}{\mathcal{L}}

\newcommand{\sT}{\mathcal{T}}
\newcommand{\sW}{\mathcal{W}}

\newcommand{\sM}{\mathcal{M}}

\usepackage{xspace}

\makeatletter
\DeclareRobustCommand\onedot{\futurelet\@let@token\@onedot}
\def\@onedot{\ifx\@let@token.\else.\null\fi\xspace}
 
\def\ie{\emph{i.e}\onedot}

\makeatother

\newcommand\paren[1]{\left(#1\right)}

\usepackage{mathtools}

\usepackage{listings}

\graphicspath{{./figs/}{./}}

\newcommand{\eat}[1]{}

\newcommand{\supp}{the Suppl. Material\xspace}

\newcommand{\ourtitle}{{When MAML Can Adapt Fast and How to Assist When It Cannot}}

\newcommand{\loss}{\ell}
\newcommand{\Hessian}{H}
\DeclareMathOperator*{\E}{\mathop{\mathbb{E}}}  

\newcommand{\mamlloss}{\sL^{\textsc{maml}}}
\newcommand{\mamlsol}{\theta^{\textsc{maml}}}

\usepackage[colorinlistoftodos, textsize=tiny, textwidth=34mm]{todonotes}
\definecolor{sicolor}{rgb}{.5, 1, .5}

\begin{document}

%

%

\twocolumn[

\aistatstitle{\ourtitle}

\aistatsauthor{S\'ebastien M. R. Arnold \And Shariq Iqbal \And Fei Sha }

\aistatsaddress{
    \texttt{seb.arnold@usc.edu} \\
    University of Southern California
    \And
    \texttt{shariqiq@usc.edu} \\
    University of Southern California
    \And
    \texttt{fsha@google.com} \\
    Google
} ]

\begin{abstract}
Model-Agnostic Meta-Learning (MAML) and its variants have achieved success in meta-learning tasks on many datasets and settings. Nonetheless, we have just started to understand and analyze how they are able to adapt fast to new tasks.  In this work, we contribute by conducting a series of empirical and theoretical studies, and discover several interesting,  previously unknown properties of the algorithm. First, we find MAML adapts better with a deep architecture even if the tasks need only a shallow one. Secondly, linear layers can be added to the output layers of a shallower model to increase the depth without altering the modelling capacity, leading to improved performance in adaptation. Alternatively, an external and separate neural network meta-optimizer can also be used to transform the gradient updates of a smaller model so as to obtain improved performances in adaptation. Drawing from these evidences, we theorize that for a deep neural network to meta-learn well, the upper layers must transform the gradients of the bottom layers as if the upper layers were an external meta-optimizer, operating on a smaller network that is composed of the bottom layers. 
\end{abstract}

\section{Introduction}\label{introduction}

Meta-learning or \emph{learning to learn} has been an appealing idea for addressing several important challenges in machine learning~\citep{Schmidhuber1987-kq,Bengio1991-yp,Vanschoren2019-vv,Finn2017-an}. In particular, learning from prior tasks but being able to adapt quickly to new tasks improves learning efficiency with fewer samples, \ie, few-shot learning~\citep{Vinyals2016-cm}.  A promising set of techniques, Model-Agnostic Meta-Learning or MAML~\citep{Finn2017-an} and its variants -- often referred as Gradient-based Meta-Learning (GBML) -- have attracted a lot of interest~\citep{Nichol2018-zp,Lee2018-pm,Grant2018-lh,Flennerhag2019-bg}.

In  GBML, the learning model is ``meta''-trained on a set of \emph{meta-training} tasks and is expected to perform well on \emph{meta-testing} (\ie, post-training-adaptation) tasks. In the phase of meta-training, the model parameters are optimized so that when applied to meta-testing tasks, a few gradient-based parameter updates lead to a significant reduction in the learning losses, a desideratum referred as ``fast adaptation''. To this end, MAML optimizes what is called MAML loss (\S\ref{sBackground}).

In this paper, we take an unexplored direction to understand how  MAML and its alike work: we investigate what types of model can meta-learn. 
Our work answers a few questions inspired by existing work. 

First, most research work in the literature focuses on deep learning models --- presumably one can posit that a sufficiently large deep learning model should be able to learn the right inductive bias to meta-learn as neural networks are universal approximators. While the argument is patently valid, our research work aims to refine it: what sense do we mean with \emph{sufficiently large}? Is there a regime where the model is not sufficiently large such that it cannot meta-learn? 

Second, the recently proposed ANIL algorithm suggests that for deep learning models, there is \emph{almost no need} to use the MAML loss to optimize the bottom layers of the neural network~\cite{Raghu2019-ff}.  This observation is closely related to multitask learning~\citep{Baxter2000-no,Caruana1997-je} but does not explain  what \emph{the special roles of the models' heads} are in ensuring the bottom layers are updated as effectively as the original MAML.

Third, preconditioning methods introduce additional parameters to control the gradients during the meta-testing to improve fast adaptation~\cite{Li2017-ts,Park2019-du,Lee2018-pm,Flennerhag2019-bg}. They assume those additional parameters, after being meta-trained, generalize to new tasks. However, those works do not explain why the original model can adapt fast \emph{without} those parameters. Moreover, if given a model that is not ``sufficiently large'' to meta-learn, how effective would those methods be? For example, imagine those methods were given the bottom layers of a deep neural network model, could they update those layers to match the performance of the original bigger neural network? 

To answer these questions, we need a way to measure how large a model is and a metric to measure how effective meta-learning is. For the former, we use the \emph{depth}, ie, the number of layers in deep models as it is one of the most frequently cited quantity to characterize the size of a model.  For the latter, we use the performance metrics (error rates or accuracies) on meta-testing tasks, a common practice in existing literature. We concentrate on few-shot learning tasks and leave other application scenarios of MAML and its alike to future study.

We use a theoretical analysis (of mathematically tractable models) to gain insights and to generate hypotheses around how meta-learning is enabled in deep learning models. We then use empirical studies to validate those hypotheses and inspire a new algorithm, dubbed \textsc{meta-kfo}, for meta-learning.

We summarize the key findings from our research.  We conclude that the \emph{depth of a deep model is important to meta-learning}. Even if a task is solvable with a shallow or linear model, a deeper model with at least one hidden layer is required for meta-learning. The reason is that the meta-learner needs to use the upper layers of a deep model to control how the bottom layers' parameters are to be updated. 

This control can be achieved in three ways. The first, which is the default and implicit strategy in existing work, is to use a sufficiently large deep neural network. The second one is to add linear layers to the output of a shallower network to increase the depth. This has the advantage that the adapted model is smaller  than the first approach as the linear layers can be absorbed into the shallower network.  The third one is to use especially designed preconditioning methods  that directly control the gradients that update the shallower network.  Those methods, including previous work~\citep{Li2017-ts,Park2019-du,Lee2018-pm} and the proposed \textsc{meta-kfo} algorithm (\S\ref{sKFO}), improve meta-learnability of shallower networks that otherwise do not meta-learn well. Moreover, the proposed algorithm and its empirical behavior yields new insights: we surmise  that in a deep neural network, \emph{the upper layers have the equivalent effect of transforming the gradients of the bottom layers as if the upper layers were an external meta-optimizer, operating on a smaller network that is composed only of the bottom layers}.

While it is plainly correct to state that the upperlayers of deep models affect the bottom layers' parameter updates (as in any gradient-based learning), our work is the first to refine this argument by pointing out this influence is crucial for enabling meta-learning. This is established through a mix of theoretical analysis (\S\ref{sTheory}) and empirical studies  (\S\ref{sEmpirical})  carefully designed to reveal how increasing the depth enables meta-learning.  The proposed \textsc{meta-kfo} algorithm is motivated by our findings and also contributes to the work on meta-learning by enabling shallower models to meta-learn and attaining state-of-the-art performance on several benchmarks.

\section{Background and Notation}
\label{sBackground}
In MAML and its many variants, we have a model whose parameters are denoted by $\theta$. We would like to optimize $\theta$ such that the resulting model can adapt to new and unseen tasks fast. We are given a set of meta-training tasks, indexed by $\tau$. To each such task, we associate a loss $\loss_\tau(\theta)$. Distinctively, MAML minimizes the expected task loss after an \emph{adaptation} phase, consisting of a few steps of gradient descent from the model's current parameters. Since we do not have access to how the target tasks are distributed, we use the expected loss over the meta-training tasks, 
\begin{equation}
    \mamlloss(\theta) = \E\nolimits_{\tau \sim p(\tau)}[\loss_\tau\left(\theta - \alpha \nabla \loss_\tau(\theta)\right)]
    \label{eMAMLLoss}
\end{equation}
where the expectation is taken with respect to the distribution of the training tasks. $p(\tau)$ is a short-hand for the distribution of the task: $p(\tau) = p(\theta_\tau) p(\vx, y; \theta_\tau)$,
for a set of conditional regression tasks where the data $(\vx, y)$ follows a distribution parameterized by $\theta_\tau$.  $\alpha$ is the  learning rate for the  adaptation phase.  The right-hand-side of eq.~(\ref{eMAMLLoss}) uses only one step gradient descent such that the aim is to adapt \emph{fast}: in one step, we would like to reduce the loss as much as possible. In practice, a few more steps are often used in the meta-training phase.  We use
\begin{equation}
   \mamlsol = \argmin_{\theta} \mamlloss(\theta)
\end{equation}
to denote the minimizer of this loss, \ie, the MAML solution. Note that it is most likely different from each $\loss_\tau$'s minimizer. If we use gradient descent during meta-training, the parameter is updated as follows:
\begin{equation}
(\textsc{meta-training})\quad \theta \leftarrow \theta - \beta \parde{\mamlloss(\theta)}{\theta}
\end{equation}
where the step size $\beta$ is called meta-update learning rate.   During the meta-testing, the MAML solution is used as an initialization for solving new tasks with regular (stochastic) gradient descent:
\begin{equation}
(\textsc{meta-testing})\quad \theta \leftarrow \theta - \alpha \parde{\loss_{\tau'}(\theta)}{\theta}
\end{equation}
where $\tau'$ denote a new task, and $\alpha$ is the adaptation learning rate.

\section{Overview of Our Approach}
\label{sApproach}

To understand the relationship between the depth and the meta-learnability, we start by creating a \emph{failure} scenario by identifying a base model and task setup where MAML fails to meta-learn. Then we employ a \emph{strategy of increasing the depth} of the base model such that it becomes meta-learnable. Finally, we  elucidate {what the increased depth achieves} and how it is related to existing methods of improving meta-learnability. 

To achieve these 3 desiderata, however, is challenging with deep learning models.  In essence, when the depth is increased, the improvement in performance metrics could be caused by several entangled factors.

To see this, consider a base model $\sM_1$ and a bigger model $\sM_2$ which has more layers and thus at least as powerful, if not strictly more. Suppose the adaptation performances (say, classification accuracy) are such $\sM_1^{\textsc{maml}} \le \sM_2^{\textsc{maml}}$.  With respect to these models' Bayes optimal performances, what we would like to have first is the following relation:
\begin{equation}
\sM_1^{\textsc{maml}} \le  \sM_2^{\textsc{maml}} \le \sM_1^{\textsc{bayes}}     \le  \sM_2^{\textsc{bayes}}
\end{equation} 
where we can  identify that the increase in performance metrics is solely due to the improved meta-learning when the depth is increased\footnote{Consider the alternative relation $\sM_1^{\textsc{maml}} \le \sM_1^{\textsc{bayes}} \le  \sM_2^{\textsc{maml}}    \le  \sM_2^{\textsc{bayes}}
$. Then the observed increase in performance  has several possible explanations:  the increased depth makes meta-learning more effective, improves the model's power in solving the tasks, or results in a combination of the both. Our design needs to rule the latter out.}.

It is hard to guarantee $\sM_2^{\textsc{maml}} \le \sM_1^{\textsc{bayes}}$ on real-world data as we do not know their true underlying distributions. However, in theoretical analysis, this can be achieved by analyzing problem settings where  the (base) model $\sM_1^{\textsc{bayes}}$ (and thus $\sM_2^{\textsc{bayes}}$ also)  achieves 100\% accuracy.  \S\ref{sTheory} follows this design thinking by applying  (correctly specified) models of linear regression and logistic regression to data.  The design also enables us to recognize ``failure mode'' of MAML when the base model $\sM_1^{\textsc{maml}}$ is significantly worse than $\sM_1^{\textsc{bayes}}$, say, at a chance level for  classification.

Secondly, to ensure $\sM_2$ does not increase the power of $\sM_1$ in solving the tasks, our design of theoretical analysis in \S\ref{sTheory} and empirical studies in \S\ref{sShallowDeep} and \S\ref{sLinNet} increases the depth by adding linear layers to the outputs of the base models. We refer this as ``LinNet'' strategy for adaptation. While linear layers are often cited for implicit regularization to improve generalization of models, \citep{Gidel2019-rk, Saxe2019-om} in our settings, there is no overfitting.
So those linear layers are indeed the only explanation for why meta-learning has been improved (not increased model representational capacity).
We refer the reader to \supp for details, including how collapsing deep models into shallow ones ruins meta-learnability.

To hypothesize how the depth facilitates meta-learning, our design goes beyond the standard argument that the upper layers of a deep neural net or added linear nets influence the bottom layers' gradients. We specifically design an algorithm called \textsc{meta-kfo} (\S\ref{sKFO}) where a separate neural network learns to \emph{explicitly} transform the gradients and enable meta-adaptation on shallower models that otherwise adapt poorly.  

The most important feature of this algorithm is to keep the base model's modeling capacity unchanged. This allows us to directly compare to deep models. Our empirical observations support the hypothesis that in a deep neural network, the upper layers transform the gradients of the bottom layers \emph{as if the bottom layers alone were being meta-trained}.

\section{Theoretical Analysis}
\label{sTheory}

We conduct theoretical analysis on mathematically tractable models and task setups, following the design outlined in the  previous section.
We start with creating a failure mode of MAML by employing a 1-D regression as a base model that is not meta-learnable but nonetheless can solve the meta-testing tasks.
We then increase the size of the base model to make it meta-learnable by overparameterzing it (\ie, adding a ``linear layer'').
Albeit unrealistic, this setup is of didactic value and also admits a simple analytical solution\footnote{Others have studied the multi-dimensional version of this setup under different perspectives~\cite{pmlr-v97-balcan19a, Saunshi2020-cu, denevi2019learning}}.
We describe the setup in \S\ref{sAnalysisSetup} and empirical observations of the base and the overparameterized models in \S\ref{sAnalysisShallowDeep}.
We analyze them in \S\ref{sAnalysisShallow2} and \S\ref{sAnalysisDeep2} and contrast the difference in parameter updates for both meta-training and meta-testing.
The insights are discussed in \S\ref{sAnalysisDiscuss}, which motivates our empirical studies in \S\ref{sEmpirical}.

\subsection{Setup}
\label{sAnalysisSetup}
 We consider the task of one-dimensional linear regression. Let the task parameter $\theta_\tau \sim N(0,1)$ be a normal distributed scalar and likewise, the covariate $x\sim N(0,1)$. The observed  outcome is $y \sim N(\theta_\tau x, 1)$. 
We investigate two models for their meta-learning performance:
\begin{align}
\textsc{shallow:}\quad  \hat{y} & = cx\\
\textsc{deep:}\quad \hat{y} & = abx
\end{align}
Note that the ``deep'' model is overparameterized and can be seen as two-layer neural nets with weights being $a$ and $b$ respectively. For each task, we use the least-square loss
\begin{align}
\loss_\tau(c) &= \E\nolimits_{p(x, y|\theta_\tau)} (y - cx)^2\\
   \loss_\tau(ab) & = \E\nolimits_{p(x, y|\theta_\tau)} (y - abx)^2
\end{align} 
Note that the data of these tasks are generated according to the models used for meta-learning.

\begin{figure}[t]
\centering
\small
    \includegraphics[width=0.22\textwidth]{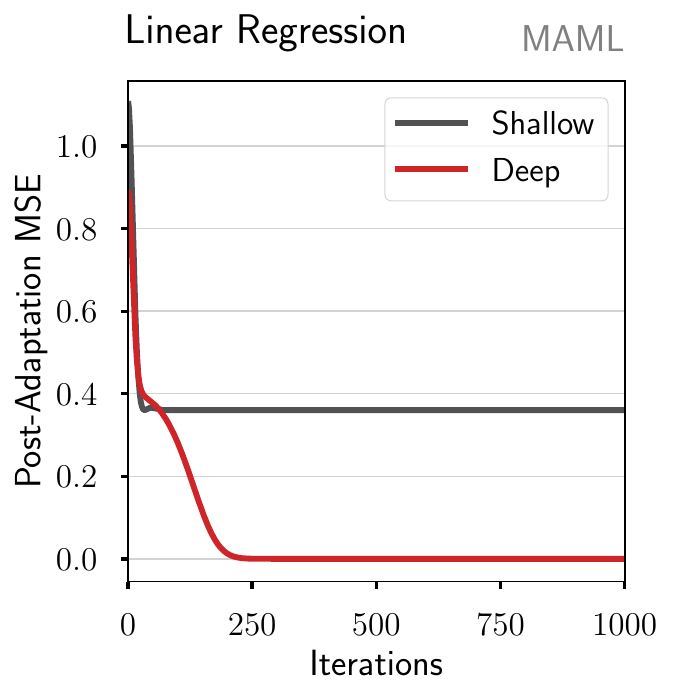}
    \includegraphics[width=0.22\textwidth]{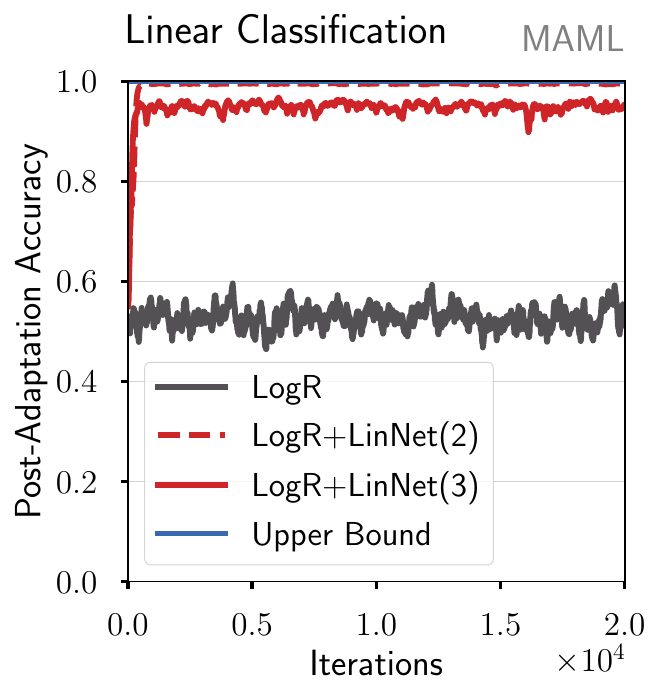}
\caption{\small \textsc{shallow} models for regression (Left) and classification (Right) fail but overparameterized \textsc{deep} models are able to meta-learn.}
\label{fShallowDeepAnalysis}
\vskip -1em
\end{figure}

\subsection{\textsc{shallow} fails; \textsc{deep} meta-learns}
\label{sAnalysisShallowDeep}
Fig.~\ref{fShallowDeepAnalysis}(left) contrasts the two models' surprising differences in performance on meta-learning. While the \textsc{deep} (the red curve) quickly reduce the MSE  on meta-testing tasks, the black curve demonstrates the poor performance of the MAML algorithm on the \textsc{shallow} model.

The results are unexpected as both models are fully capably of solving the problem given enough data --- in particular, the \textsc{shallow} model has  only one parameter $c$ to learn. 

In Fig.~\ref{fShallowDeepAnalysis}(right), we  show a similar study of meta-learning linear  classifiers where the data is generated according to the models. The base model is $\hat{y} = \textsc{bernoulli}(\sigma(c\T x))$ and its overparameterized version  $\hat{y} = \textsc{bernoulli}(\sigma(a\T Bx))$ where $a, B$ and $c$ are matrices or vectors and $x$ is a vector. (For details, see the \supp). The base model attains an accuracy at the chance level while the overparameterized one reaches near-perfect classification accuracy. As in the 1-D regression, the overparameterization (equivalent to adding two or three linear layers) enables meta-learning and fast adaptation. \emph{What roles could those additional parameters  have played}?

\subsection{Analysis of the \textsc{shallow} model}
\label{sAnalysisShallow2}

\paragraph{The MAML Solution} It is easy to see that the MAML solution is the origin $c^{\textsc{maml}}=0$ given the symmetries in both $x$ and $\theta_\tau$. We state the following results (see \supp for proofs):
\begin{align}
\loss_\tau(c - \alpha\nabla\loss_\tau) & = (1-\alpha)^2(c-\theta_\tau)^2 +\textsc{const}\\
 \mamlloss_{\textsc{shallow}} & = 2(1-\alpha)^2c^2 +\textsc{const}
\end{align}
where the MAML loss is a convex function with the minimizer at $c^{\textsc{maml}} = 0$, in accordance with our intuition. The gradient of the $\tau$th task is given by
\begin{equation}
\parde{\loss_\tau(c - \alpha\nabla\loss_\tau) }{c} \propto (1-\alpha)^2 (c-\theta_\tau)
\label{eShallowGrad}
\end{equation}
Note the the gradient is proportional to the deviation from the ``ground-truth'' parameter $\theta_\tau$. The  parameter updates during  meta-training and adaptation  are given by (cf. \S\ref{sBackground}))
\begin{align}
(\textsc{meta-training})\quad c & \leftarrow c - \beta (1-\alpha)^2 (c-\theta_\tau)
\label{eShallowUpdateTrain}\\
(\textsc{meta-testing})\quad  c & \leftarrow c -  \alpha  (c-\theta_\tau)
\label{eShallowUpdateTest}
\end{align}

\paragraph{No One Step Adaptation} Suppose we would like to adapt from a task whose parameter is $\theta'$, from the MAML solution $c^{\textsc{maml}} = 0$, we get
\begin{equation}
c \leftarrow c^{\textsc{maml}} - \alpha(c^{\textsc{maml}} - \theta') = \alpha \theta'
\end{equation}
Thus, unless $\alpha$ happens to be 1, the optimal solution cannot be achieved in one step of adaptation. However, when $\alpha =1$, the gradient of $\mamlloss_{\textsc{shallow}}$ is zero (cf. eq~(\ref{eShallowGrad})),  thus meta-learning cannot occur.

\subsection{Analysis of the \textsc{deep} model}
\label{sAnalysisDeep2}

\begin{figure}[t]
    \centering
    \includegraphics[width=0.23\textwidth]{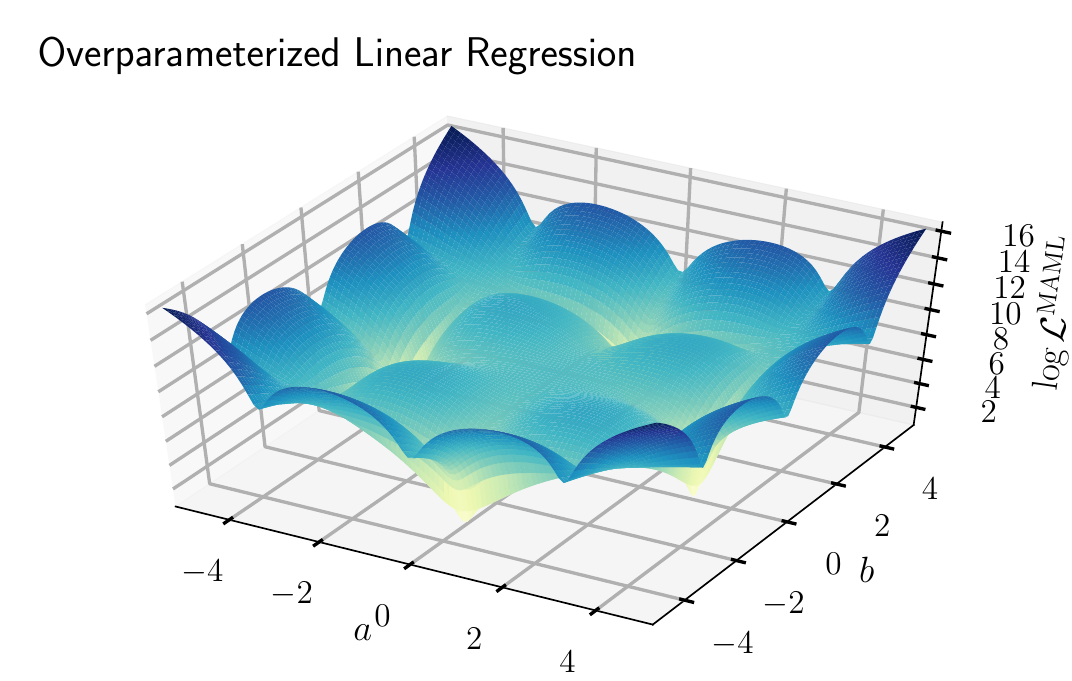}
    \includegraphics[width=0.23\textwidth]{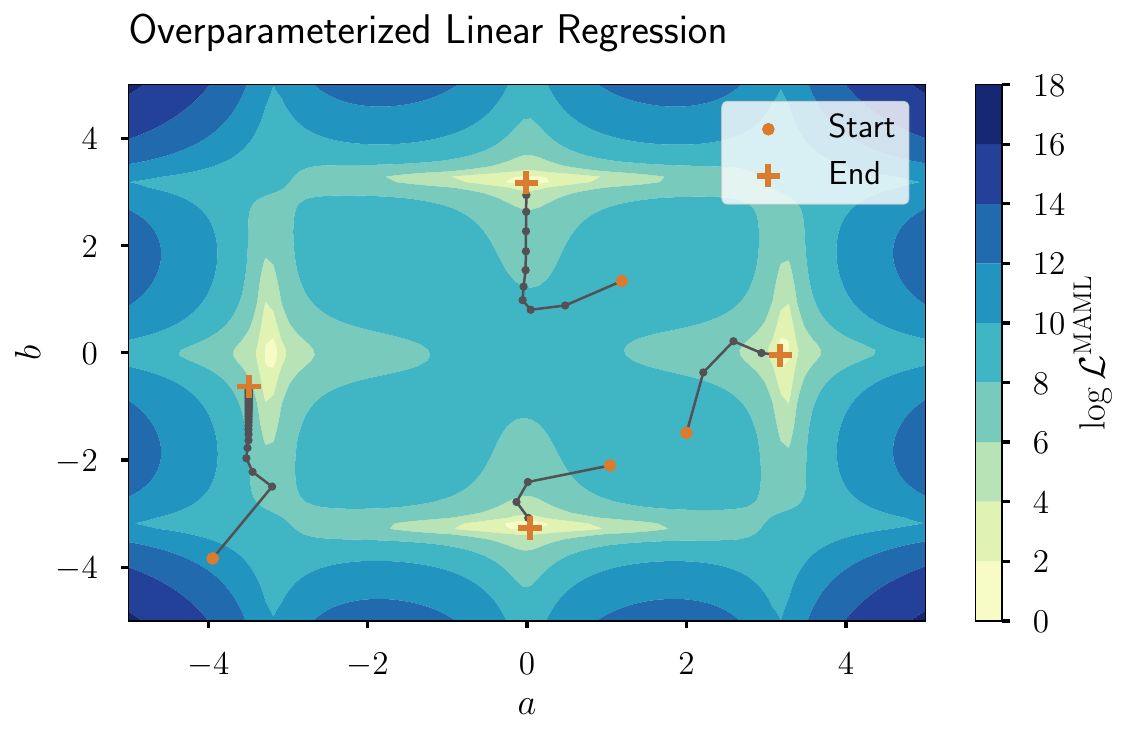}
    \caption{\small  Meta-learning of a 1D linear regression model (\S\ref{sAnalysisSetup}). \textbf{(Left)} MAML loss of \textsc{deep}, showing multiple (local) minima with deep valleys.  \textbf{(Right)} 4 meta-training trajectories (of parameters) converging to each of the 4 solutions.     
        } \label{fLinearAnalysis}
    \vspace{-1em}
\end{figure}

\paragraph{The MAML solution}
Unfortunately, for the \textsc{deep} model, both the gradients and the losses are very complicated.  With the details in \supp, we state the following

\begin{itemize}
\item The origin of the parameter space $(a=0, b=0)$ is a stationary point and the Hessian  is $\Hessian = -4\alpha\mI$. Thus,  the origin is a local \emph{maximum}, thus \emph{not} a MAML solution.
\item The following 4 pairs of $(a^{\textsc{maml}}= \pm 1/\sqrt{\alpha}, b^{\textsc{maml}} =0)$ and $(a^{\textsc{maml}}= 0, b^{\textsc{maml}}=\pm 1/\sqrt{\alpha})$ are locally \emph{minimum}, with the Hessian given by $\textsc{diag}(8\alpha, 6\alpha^3)$, are thus MAML solutions.
\end{itemize}
We visualize $\mamlloss_{\textsc{deep}}$ in Fig.~\ref{fLinearAnalysis}, where we can see clearly the 4 local minimum (as well as the deep valleys, ``ravines'') and how trajectories of parameter updates converge to them.

The gradients involve high-order polynomials of $a$ and $b$ -- they are given in the \supp. To gain insights, in the below,  we hold $a$ fixed and examine the gradient with respect to $b$ during meta-training. This is reminiscent of ANIL~\cite{Raghu2019-ff}. The resulting form of the gradients is greatly simplified yet remains insightful:
\begin{equation}
\parde{\loss_\tau(a(b - \alpha\nabla_b\loss_\tau)) }{b} \propto a(1-\alpha a^2)^2 (ab-\theta_\tau)
\end{equation}
Note that the symbol $\nabla_b$ indicate that only $b$ is meta-learned with $a$ fixed. This leads to the following:
\begin{align}
(\textsc{meta-training})\ b &\leftarrow b - \beta a(1-\alpha a^2)^2 (ab-\theta_\tau)\label{eDeepUpdateTrain}\\
(\textsc{meta-testing})\ b & \leftarrow b - \alpha a (ab-\theta_\tau) \label{eDeepUpdateTest}\\
\ a & \leftarrow a - \alpha b (ab-\theta_\tau)
\end{align}

\paragraph{One Step Fast Adaptation} The \textsc{deep} model has qualitatively very different adaptation behavior from the \textsc{shallow} model. As before,  at the MAML solution $(a^{\textsc{maml}} = 1/\sqrt{\alpha}, b^{\textsc{maml}}=0)$, we perform an adaptation on a new task with the ground-truth $\theta'$.  Holding $a^{\textsc{maml}}=1/\sqrt{\alpha}$ fixed,  the update to $b$ is
\begin{equation}
b^{\textsc{new}} \leftarrow b^{\textsc{maml}} - \alpha a^{\textsc{maml}} (a^{\textsc{maml}}b^{\textsc{maml}} - \theta') = \sqrt{\alpha} \theta'
\label{eLinearDeepFast}
\end{equation}
Note that $(a^{\textsc{maml}} = 1/\sqrt{\alpha}, b^{\textsc{new}} = \sqrt{\alpha} \theta')$ is \textbf{precisely} the optimum solution to the task as $a^{\textsc{maml}}b^{\textsc{new}} = \theta'$. In other words, we need only one parameter update to arrive at the optimum solution!   In fact, this \textbf{fast} adaptation does \textbf{not} depend on what $\alpha$ is and does not even depend on whether we adapt from the right $b^{\textsc{maml}}$ --- for any random $b^{\textsc{random}}$, the update in eq.~(\ref{eLinearDeepFast}) immediately brings $b^{\textsc{random}}$ to $b^{\textsc{new}}=\sqrt{\alpha}\theta'$!

\subsection{Insights from \textsc{shallow} versus \textsc{deep}}
\label{sAnalysisDiscuss}

We examine the gradient updates of the two models. First, for adaptation during meta-testing, both eq.~(\ref{eDeepUpdateTest})  and eq.~(\ref{eShallowUpdateTest})  share the same element being proportional to the error signal: $(ab-\theta_\tau)$ for \textsc{deep} and $(c- \theta_\tau)$ for \textsc{shallow}.  However the additional factor $a$ in the \textsc{deep} model enables \textbf{one-step fast adaptation} that is not possible to attain by the \textsc{shallow model}, as shown in the previous section.

Turning to the meta-training, we also notice the different scaling factors to the error signals. Contrasting  eq.~(\ref{eDeepUpdateTrain}) to eq.~(\ref{eShallowUpdateTrain}), the effective step size for the former depends on $a(1-\alpha a^2)^2$ and cannot be absorbed into the meta-learning rate $\beta$ if $a$ is also updated. Namely, in the meta-training phase, the step size for updating model parameters is \textbf{dynamically adjusted}, while the step size for the \textsc{shallow} model is \textbf{fixed}.

While fully characterizing how $a$'s  dynamics change parameter updates is left for future work, we concentrate our analysis in the neighborhood of the solutions, say,  $(a^{\textsc{maml}} = 1/\sqrt{\alpha}, b^{\textsc{maml}} = 0 )$ (the other 3 are symmetric to this one).  Note that the farther $a$ is away from $a^{\textsc{maml}}$, the bigger the step size (in magnitude) is to amplify the error signal $(ab - \theta_\tau)$. This has the effect to move $b$ more quickly toward the solution $\theta_\tau/a$ for the $\tau$-th task, or toward the MAML solution $b^{\textsc{maml}} = 0$ (as the expectation with respect to the task distribution is 0). 

Furthermore, when at the MAML solution $(a^{\textsc{maml}}=1/\sqrt{\alpha},b^{\textsc{maml}} = 0 )$,  the update eq.~(\ref{eDeepUpdateTrain}) is stationary for \emph{any} task. Imagine a new task $\tau'$ with ground-truth parameter $\theta'$ is randomly sampled for meta-training.  Even if $a^{\textsc{maml}}b^{\textsc{maml}} \neq \theta'$, the gradient-based update will not change $b^{\textsc{maml}}$ from $0$.
On the other hand, for the \textsc{shallow} model, even if the model is at the solution $c^{\textsc{maml}}=0$, the randomly sampled task will update the solution to $c \leftarrow - \alpha \theta'$, drifting away from the MAML solution.  In other words, the MAML solution is more stable in \textsc{deep} than in \textsc{shallow}, when (as a common practice) stochastic gradient descent is used.

We leave more comprehensive characterization of the local and global dynamics of the parameter updates to future work. In this work, our focus is:  \emph{how these observations shed light on more complicated models used in  practice}?

The first insight is that even with the same modeling capacity, depth plays an important role in enabling meta-learning, even if the depth is the form of an additional scalar parameter or additional linear layers. 

The second insight comes from extending the 1-D linear regression to multi-dimensional regression where the base model is $\hat{y} = Cx$ and its overparameterized version as $\hat{y}  = a\T Bx$. It is not hard to see the forms of the gradients suggest that the additional parameters affect not only scaling factors (\ie, magnitude) but also \textbf{\emph{transforming  gradient directions through those additional parameters}}. While hard to analyze mathematically, our empirical studies below provide strong evidences.

\begin{figure}[t]
    \centering
    \small
   \includegraphics[width=0.23\textwidth]{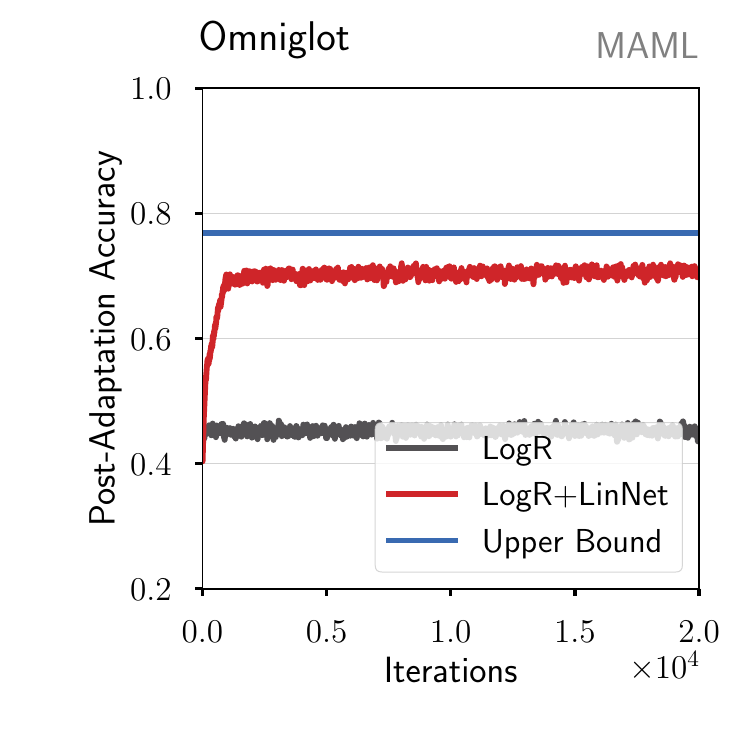}
   \includegraphics[width=0.23\textwidth]{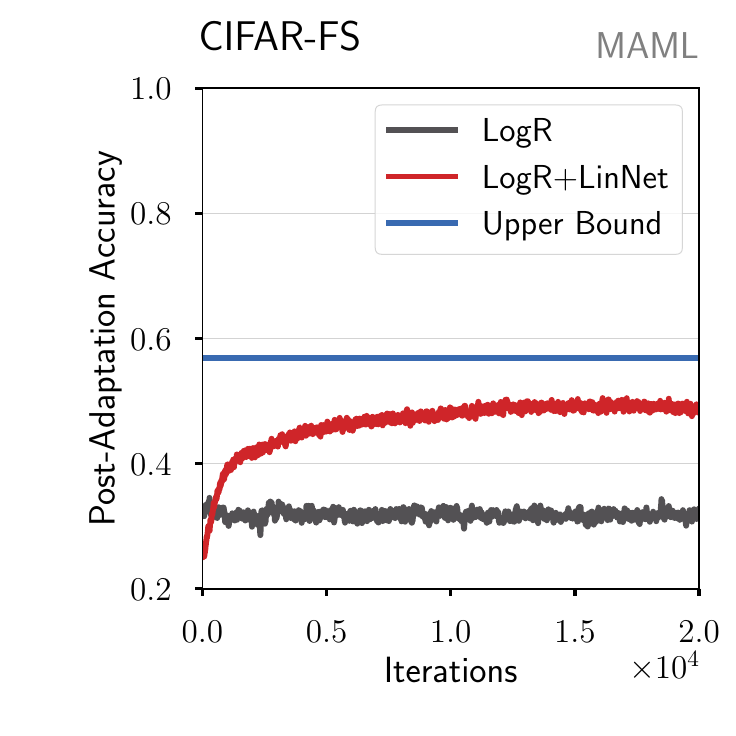}
    \vspace{-1em}
      \caption{
          \small
          Meta-training logistic regression models with MAML on Omniglot and CIFAR-FS led to poor performances (mini-ImageNet results are in \supp).
          Adding linear nets improves meta-learning significantly, \emph{without} changing the model's capacity.}
      \label{fLogRegLinNet}
      \vspace{-2.0em}
\end{figure}

\section{How to Be (More) Meta-Learnable} \label{sEmpirical}

The analysis on highly idealized models in the previous section needs to be empirically verified on real-world datasets. We first validate the findings in \S\ref{sAnalysisShallowDeep} by showing that adding linear layers (``LinNet'') as a general strategy of increasing the depth of the models improves meta-learning in both shallow models (\S\ref{sShallowDeep}) and deep models (\S\ref{sLinNet}). To further clarify the roles of the upper layers in deep models for meta-learning,  we propose a new algorithm for meta-learning in \S\ref{sKFO},  and conduct additional empirical studies. While this algorithm \textsc{meta-kfo} is primarily used in this work for investigating how MAML works, it also attains state-of-the-art performances.  We perform our study on  common benchmarks datasets of meta-learning (for few-shot classification).

\paragraph{Datasets and settings}
In the following study, we use the standard 5-ways and 5-shots setting on the Omniglot \citep{Lake2015-vz}, CIFAR-FS \citep{Bertinetto2018-gn}, and mini-ImageNet \citep{Vinyals2016-cm} datasets.
We denote by CNN($X$) the convolutional network with $X$ convolutional layer; for example, CNN(4) corresponds to the baseline network also used in \citet{Finn2017-an} and \citet{Raghu2019-ff} among many others.
To ensure fair comparison, we independently reimplemented each algorithmic variant and found the best hyper-parameter values for each architecture-algorithm pair via grid-search.
For additional details, see \supp.

\subsection{Linear layers improve shallow linear models}
\label{sShallowDeep}

Fig.~\ref{fLogRegLinNet} displays the results of meta-learning using MAML on two standard benchmark datasets  with logistic/softmax regression  models. The light blue horizontal lines denote the best performance if the models are trained by sufficient data from the meta-testing tasks.  The black lines are the meta-learning performance, which only slightly improve upon chance levels (20\% on Ominglot and CIFAR-FS). 

However, as in Fig.~\ref{fShallowDeepAnalysis}, when linear layers are added to these linear models, the meta-learning performances are significantly improved. 

\subsection{Linear layers improve deep nonlinear models}
\label{sLinNet}

Table~\ref{tLinNetMAML} lists the positive results of adding two linear layers to  different CNN architectures.
When the number of CNN layers are less than 6, the addition improves meta-learning performances.
At CNN(6), there are degradations in performance by the original MAML on Omniglot and CIFAR-FS such that LinNet does not improve further.
On mini-ImageNet, while MAML improves, LinNet decreases though its performance at CNN(4) is still the best. 
When degradation occurs, it is marginal ($-0.4\%$ for CNN4 on Omniglot and $-0.5\%$ for CNN6 on mini-ImageNet) and not alarming: for Omniglot, the standard deviation on accuracies is $\pm 0.76\%$ and on mini-ImageNet it is $\pm 1.12\%$.

\begin{table}[t]
\vspace{-0em}
\centering
\setlength{\tabcolsep}{3pt}
\caption{\small Accuracy Improves by Adding Linear Layers}
\label{tLinNetMAML}
\vspace{-1em}
\small
\begin{tabular}{l cccc |cccc }
\addlinespace
\toprule
Method        &  \multicolumn{4}{c}{MAML} & \multicolumn{4}{c}{MAML w/ LinNet} \\
\midrule
CNN Layers    &2 & 3    & 4    & 6    & 2  & 3    & 4    & 6   \\ 
\midrule
Omniglot      &66.8 & 93.5 & 98.5 & 97.6 & 88.1 & 95.5 & 98.1 & 97.6  \\
CIFAR-FS      &62.2 & 68.9 & 70.9 & 71.3 &66.1 &  71.1 & 74.4 & 71.9 \\
mini-ImageNet &52.6 & 54.0 & 64.1 & 64.6 & 60.5 &  60.2 & 64.9 & 64.1 \\
\bottomrule
\end{tabular}
\vspace{-0.0em}
\end{table}

Table~\ref{tLinNetANIL} generalizes the positive findings in Table~\ref{tLinNetMAML} to ANIL~\citep{Raghu2019-ff}.
Thus, we believe LinNet is a broadly applicable strategy for improving  meta-learning.

\begin{table}[t]
\vspace{0em}
\centering
\caption{\small Accuracy Improves on ANIL Trained CNN(2)}
\label{tLinNetANIL}
\vspace{-1em}
\setlength{\tabcolsep}{3pt}
{\small
\begin{tabular}{l cc cc}
\addlinespace
\toprule
Dataset         & w/o LinNet & w/ LinNet \\
\midrule
Omniglot      &  91.00  & 93.02  \\
CIFAR-FS      &   66.10  & 67.55    \\
mini-ImageNet &   56.42  & 56.64  \\
\bottomrule
\end{tabular}
}
\vspace{-0em} 
\end{table}

\subsection{Meta-Optimizer for fast adaptation}
\label{sKFO}

\paragraph{Main idea} It is straightforward to see that the added linear layers (LinNets)  function similarly to the upper layers of deep learning models. The parameter updates for the bottom layers before such layers are modulated by the parameters in the upper layers or the LinNets. However, in what ways does this modulation help meta-learning?

Related to this question is meta-learning via learning to optimize, \ie, transforming the gradients of the models~\cite{Li2017-ts,Park2019-du,Lee2018-pm,Flennerhag2019-bg}. Those types of preconditioning techniques could also be used to make a (smaller) model (more) meta-learnable. Thus, are the parameter updates in deep models equivalent  to transformed gradient updates by such techniques? Note that there is a subtle difference: in some of these techniques (such as T-Nets and WarpGrad), the loss function used to compute the gradients with respect to the bottom layers \emph{prior to} transformation actually contains the the transformation parameters themselves, cf. eq.(\ref{eTNets}) for an example. This type of ``inline'' transformations de facto increase the model capacity by injecting more parameters. 

Our goal, however, is different and aims to disentangle the increase in model capacity from the ability to transform gradients.  The empirical observation of this approach will enable us to answer the aforementioned question more clearly. 

In the following we give a brief account of various approaches for learning to optimize and our proposed \textsc{meta-kfo} algorithm. The details are in the \supp.  \textsc{meta-kfo} is able to merely transforming the gradients of a smaller model without increasing its modeling capacity but still results in better meta-learnability. Furthermore, the improvement diminishes when the smaller model gets bigger. We surmise why sufficiently large deep models can meta-learn: \emph{the upper layers have the equivalent effect of transforming the gradients of the bottom layers as if the upper layers were an external meta-optimizer, operating on a smaller network that is composed only of the bottom layers}.

\paragraph{{\textsc{meta-kfo}} and other meta-optimizion methods}
A meta-optimizer  is a parameterized function $U_\xi$ defining the model's parameter updates. For example, a linear meta-optimizer might be defined as:
\begin{equation}
    U_\xi(g) = A g + b,
\end{equation}
where $\xi = (A, b)$ is the set of parameters of the linear transformation.
The objective is to jointly learn the model and optimizer's parameters $\xi$ to accelerate optimization. Motivated by the analysis of meta-learning in deep nets, we propose to use such an optimizer to transform the gradient updates:
\begin{equation}
\mamlloss_{\textsc{mo}}(\theta)= \E\nolimits_{\tau \sim p(\tau)}[\loss_\tau\left(\theta - \alpha U_\xi( \nabla \loss_\tau(\theta))\right)]
\end{equation}
that takes a similar role of the upper-layers in deep nets in minimizing the MAML loss:
\begin{equation}
\theta \leftarrow \theta - \beta \parde{\mamlloss_{\textsc{mo}}(\theta)}{\theta},\qquad \xi \leftarrow \xi - \beta \parde{\mamlloss_{\textsc{mo}}(\theta)}{\xi}
\end{equation} 
where $\beta$ is the meta-update learning rate.
In this notation, Meta-Curvature~\cite{Park2019-du} implements
\begin{equation}
\textsc{(mc)}\quad U_{\xi}\paren{\nabla \loss\paren{\theta}} = M \nabla \loss\paren{\theta}
\label{eMC}
\end{equation}
where $M$ is a matrix ( block-diagnonal tensor factorized). When M is diagonal, this becomes the Meta-SGD~\cite{Li2017-ts}. Furthermore, when M is identity, this become MAML.  For T-Nets (to be used with MAML-loss), the model parameters are expanded with affine transformations,
\begin{equation}
\textsc{(t-nets)}\quad \loss_\tau\left(\sA({\theta}) - \alpha \nabla \loss_\tau(\sA({\theta})))\right)
\label{eTNets}
\end{equation}
where the transformation $\sA(\cdot)$ contains two components $(\sW, \sT)$. $\sT$ is shared by all the tasks and $\sW$ is task-specific.
Since $\sA$ is linear, it can be absorbed into the original model after adaptation.  For WarpGrad \citep{Flennerhag2019-bg}, the transformation $\sA$ is defined with nonlinear layers, thus strictly increasing the size of the original model (thus, is not considered in this work).

Our method takes the form
\begin{equation}
\textsc{(meta-kfo)}\quad U_{\xi}\paren{\nabla \loss\paren{\theta}} = f( \nabla\loss\paren{\theta}; \phi)
\end{equation}
where $f(\cdot)$ is a nonlinear function parameterized by a set of parameters $\phi$ that is \emph{independent} of the model's parameters $\theta$. This approach generalizes MC (eq.~(\ref{eMC})), as it is more adaptive since the gradient $\nabla\loss\paren{\theta}$ is used as the inputs.

For models with a large number of parameters, the transformation $U$ (ie, $\sA$, $M$, and $f(\cdot)$) could contain a lot of parameters and incur high computational cost. For details, please refer to the cited references, and the \supp on the details of \textsc{meta-kfo}. Essentially, $f(\cdot)$ is parameterized with a neural network where the gradients $\nabla\loss\paren{\theta}$ are manipluated with Kronecker products.

\begin{figure}[h]
    \begin{center}
            \includegraphics[width=0.45\columnwidth]{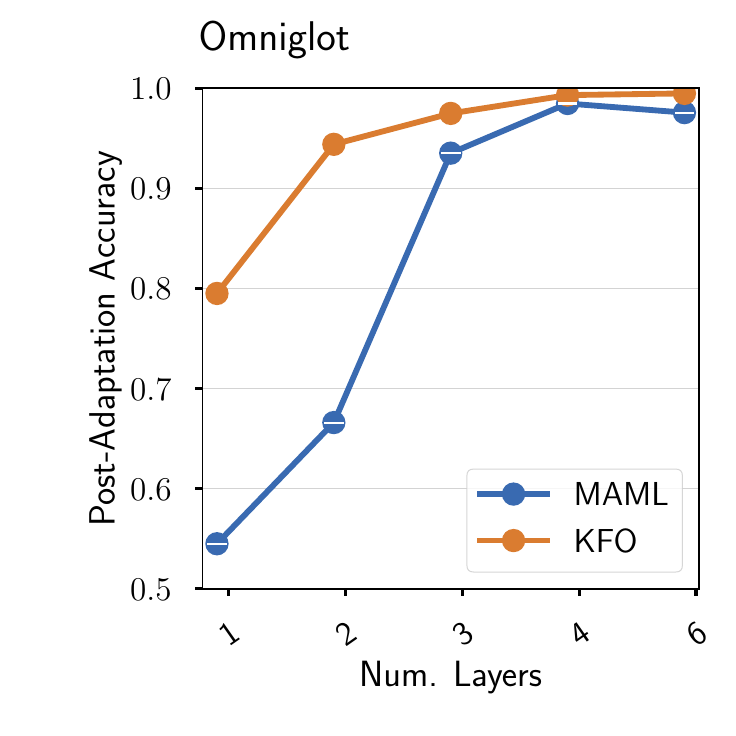}      
            \includegraphics[width=0.45\columnwidth]{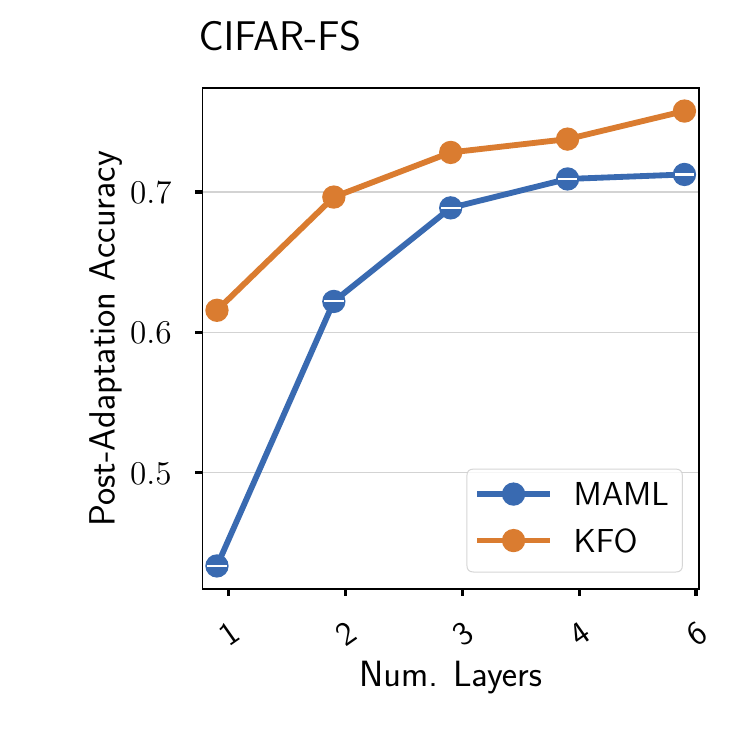}
    \vspace{-1em}
    \end{center}
    \caption{\small 
    The effect of the number of convolutional layers on adaptation performance.
    First, as the model size increases, the performances of both methods improve.
    Besides better meta-learning, the improvement can also be caused by the model's increased capacity to learn the target tasks.
    Secondly, the ``net gain'' from the \textsc{meta-kfo} has the diminishing trend as the size increases.
    In other words, the benefits of directly transforming gradients with an external meta-optimizer reduce as the upper layers of the larger models have more capacity to meta-learn to control their own bottom layers.
    Results on mini-ImageNet are available in \supp.}
    \label{fig:exp-layers}
    \vspace{-1em}
\end{figure}

\begin{table}[t]
\vspace{0em}
\centering
\caption{\small Meta-Optimizers Outperform MAML on CNN(2)}
\vspace{-1em}
\label{tShallowAdaptationMetaOptimizaer}
\setlength{\tabcolsep}{3pt}
{\small
\begin{tabular}{l ccccc}
\addlinespace
\toprule
Dataset  & MAML &   \multicolumn{4}{c}{MAML w/} \\
\cmidrule{3-6}

              &      & MSGD           & MC    & T-Nets & \textsc{meta-kfo}            \\
\midrule
Omniglot      & 66.6 & 74.07          & 94.63 & 92.27  & \textbf{96.62}  \\
CIFAR-FS      & 62.2 & 62.82          & 68.37 & 66.42  & \textbf{69.64} \\
mini-ImageNet & 52.6 & \textbf{59.90} & 58.95 & 58.47  & 59.08 \\

\bottomrule
\end{tabular}
}
\vspace{-1em}
\end{table}

\paragraph{Results} Table~\ref{tShallowAdaptationMetaOptimizaer} contrasts  different approaches for improving meta-learning by MAML on CNN(2) \emph{without} increasing the size of the model after adaptation. All methods improve the original MAML while \textsc{meta-kfo} improves the most on Omniglot and CIFAR-FS.
On mini-ImageNet, all methods improve about the same amount.
In \supp \S B.5, we also contrast the meta-optimizers when ANIL is used to meta-learn.
Again, all methods improve the baseline ANIL and \textsc{meta-kfo} improves the most significant.
 
Fig.~\ref{fig:exp-layers} examines the improvement of \textsc{meta-kfo} over MAML with respect to the network size.
As expected, \textsc{meta-kfo} improves the most when the model is small and the improvement reduces  when the model is sufficiently large.
In other words, when the model is deep enough to meta-learn by itself using its top-layers to control the gradients of bottom layers, there is less advantage of using an external meta-optimizer to learn the bottom layers.  

We view this as a strong evidence to support the theory that for deep neural networks that can meta-learn well, the upper layers have the equivalent effect of transforming the gradients of the bottom layers as if the upper layers were an external meta-optimizer, operating on a smaller network that is composed of the bottom layers.
\emph{They are the ``external meta-optimizers that work from the inside.''}.

\section{Related Work}\label{sec:related-work}

Understanding how MAML and its alike work continues to draw research interests~\citep{Finn2017-uj,Fallah2019-rc,Raghu2019-ff,Saunshi2020-cu}.
Many such studies have left open questions to be carefully analyzed, and hypotheses to be tested.

\citet{Finn2017-uj} showed that, when combined with deep architectures, GBML is able to approximate arbitrary meta-learning schemes.
That work would have assumed the model is meta-learnable to begin with, relying on the argument that deep models are universal approximators.
\citet{Fallah2019-rc} provided convergence guarantees for MAML.
Other analyses have attempted to explain the generalization ability of GBML~\citep{Guiroy2019-pa, Nichol2018-zp}, the bias induced by restricting the number of adaptation steps~\citep{Wu2018-cf}, or the effect of higher-order terms in the meta-gradient estimation~\citep{Foerster2018-qy,Rothfuss2018-lm}.
Those work do not directly investigate what elements in deep models make them meta-learn well.

\cite{Raghu2019-ff} suggested that the bottom layers of a neural network learn representations while the upper layers are responsible for the inductive bias to adapt fast.
This observation echoes the success of some other approaches for meta-learning~\citep{snell2017prototypical, lee2019meta}.
But that work does not explain what is in the ``magic'' of the top layers to enable meta-learning.

We also investigate how adaptation could be provided by a meta-optimizer.
Meta-SGD meta-learns per-parameter learning rates~\citep{Li2017-ts} while Alpha MAML adapts those learning rates during fast adaptation~\citep{Behl2019-qc}.
Meta-Curvature learns a block-diagonal pre-conditioning matrix to compute fast-adaptation updates~\citep{Park2019-du} and T-Nets extends that by decomposing all weight matrices of the model in two separate components~\citep{Lee2018-pm}.
WarpGrad further extends T-Nets by allowing both components to be non-linear functions~\citep{Flennerhag2019-bg}.

The most salient difference of our work from existing ones is our focus on studying \emph{what makes deep models meta-learnable}.
Not only do we conclude being sufficiently deep is essential for meta-learning to succeed but we also theorize that the upper layers in the deep models essentially function as ``embedded meta-optimizers''.
Our extensive empirical studies complement the theoretical work in ~\citet{Saunshi2020-cu} which suggests that deep models might attain a lower loss than shallow ones.

\section{Conclusion}
Are deep architectures necessary for meta-learning, even if the tasks can be solved with shallow (linear) networks?
Our analysis suggests so.
How does the depth benefit meta-learning?
Our studies theorize that the upper layers of deep models learn to transform the gradients of a smaller network composed of only the bottom layers.
Thus, appending a few linear layers to a shallower network is simple yet surprisingly effective way to boosts its ability to adapt.
A more powerful but more involved one is to resort to external meta-optimizers.
We hope our observations can inspire future algorithms and studies.

\subsubsection*{Acknowledgements}
Fei Sha is on leave from University of Southern California.
We appreciate the feedback from the reviewers. This work is partially supported by NSF Awards IIS-1513966/ 1632803/1833137, CCF-1139148, DARPA Award\#: FA8750-18-2-0117,  FA8750-19-1-0504, DARPA-D3M - Award UCB-00009528, Google Research Awards, gifts from Facebook and Netflix, and ARO\# W911NF-12-1-0241 and W911NF-15-1-0484.

\clearpage

\bibliographystyle{apalike}
\bibliography{main_aistats2021}
\clearpage
\pagebreak

\onecolumn
\appendix
\setcounter{figure}{0}
\renewcommand\thefigure{\thesection.\arabic{figure}} 
\renewcommand\thetable{\thesection.\arabic{table}}
\renewcommand \thepart{}
\renewcommand \partname{}

\part{\hfill \textsc{Appendix} \hfill} 

\section{Theoretical Analysis of Linear Models for Meta-Learning}

This section provides more details to \S4 in the main text.

\subsection{Analytic Solution of 1D Linear Regression}

We use 
\begin{equation}
\theta_\alpha = \theta - \alpha \nabla \loss_\tau(\theta)
\end{equation}
as a short-hand.  The gradient of the MAML loss $\mamlloss$ is given by
\begin{equation}
    \parde{\mamlloss}{\theta} = \E\nolimits_\tau(\mI- \alpha\Hessian_\tau(\theta)) \left. \parde{\loss_\tau}{\theta}\right|_{\theta - \alpha \nabla \loss_\tau}  = \E\nolimits_\tau(I- \alpha\Hessian_\tau(\theta)) \nabla\loss_\tau (\theta_\alpha)
    \label{eMAMLGrad}
\end{equation}

We setup the problem as follows. Let the task parameter $\theta_\tau \sim N(0,1)$ be a normal distributed scalar. Let the covariate $x\sim N(0,1)$ be a normal distributed scalar and the outcome be $y \sim N(\theta_\tau x, 1)$. We investigate two models for their meta-learning performance:
\begin{equation}
\textsc{shallow:}\quad  \hat{y} = cx, \quad \textsc{deep:}\quad \hat{y} = abx
\end{equation}
Note that the ``deep'' model is overparameterized. For each task, we use the least-square loss
\begin{equation}
\loss_\tau = \frac{1}{2}\E\nolimits_{p(x, y|\theta_\tau)} (y - cx)^2, \quad \text{or}\quad  \loss_\tau = \frac{1}{2}\E\nolimits_{p(x, y|\theta_\tau)} (y - abx)^2
\end{equation} 

\subsubsection{Shallow Model}
For the shallow model, the gradient is
\begin{equation}
\nabla\loss_\tau = \E\nolimits_{p(x, y|\theta_\tau)} (cx - y)x = c - \theta_\tau
\end{equation}
Thus, the loss for the $\tau$-th task is given by
\begin{align}
\loss_\tau(c - \alpha\nabla\loss_\tau) & = \E\nolimits_{p(x, y|\theta_\tau)} ( y - [(1-\alpha)c +\alpha\theta_\tau]x)^2\\
& = [(1-\alpha)c + \alpha\theta_\tau]^2 - 2 [(1-\alpha)c+\alpha\theta_\tau]\theta_\tau
\end{align}
Simplifying and completing the square on the $\theta_\tau^2$ term with constants, we arrive at
\begin{align}
\loss_\tau(c - \alpha\nabla\loss_\tau)  & = (1-\alpha)^2(c-\theta_\tau)^2 +\textsc{const}
\end{align}
Taking the expectation of the loss with respect to $p(\tau)$, we have
\begin{align}
 \mamlloss_{\textsc{shallow}}& = 2(1-\alpha)^2c^2 +\textsc{const}
\end{align}

\subsubsection{Deep Model}

The gradients of  the linear regression loss w.r.t $a, b$ are given by:

\begin{align}
    \frac{\partial l_\tau}{\partial a} &= ab^2 - b\theta_\tau \\
    \frac{\partial l_\tau}{\partial b} &= a^2b - a\theta_\tau
\end{align}

The post-adaptation parameters are obtained by taking one step of gradient descent with learning rate $\alpha$:
\begin{align}
    a \gets a - \alpha b \paren{ab - \theta_\tau} \\
    b \gets b - \alpha a \paren{ab - \theta_\tau}
\end{align}
This gives us the expression for the MAML loss:
\begin{align}
    \label{eq:maml_loss}
    \mathcal{L}^\text{MAML}_{\textsc{deep}}
    &= \frac{1}{2}\E_\tau \E_{x, y}  \left[ y - (a - \alpha b(ab - \theta_\tau)) \cdot (b - \alpha a(ab - \theta_\tau))x \right]^2 \\
    &= \frac{1}{2}\E_\tau \E_{x, y}   \left[ y - f(a, b, \theta_\tau)x \right]^2 \\
    &= \frac{1}{2}\E_\tau   \left[ \theta_\tau^2 -2 \theta_\tau + f^2(a, b, \theta_\tau) \right] \\
    &= \frac{1}{2} + \frac{1}{2} \E_\tau f^2(a, b, \theta_\tau) - \frac{1}{2} \E_\tau \theta_\tau f(a, b, \theta_\tau)
\end{align}
where $f(a, b, \theta_\tau) = (a - \alpha b(ab - \theta_\tau)) \cdot (b - \alpha a(ab - \theta_\tau))$.
\newline
We now take a closer look at the term $\E f^2$:
\begin{align}
     f^2(a, b, \theta_\tau)
    &=   \left[ ab - \alpha a^2 (ab - \theta_\tau) - \alpha b^2 (ab - \theta_\tau) + \alpha^2ab(ab - \theta_\tau)^2 \right]^2 \\
    &=   \left[ ab - \alpha(a^2 + b^2)(ab - \theta_\tau) + \alpha^2 (ab - \theta_\tau)^2 \right]^2 \\
    &=   \left[ ab - \alpha (a^2 + b^2)ab + \alpha^2 a^3 b^3 + (\alpha (a^2 + b^2) - 2\alpha^2 a^2 b^2) \theta_\tau + \alpha^2 ab\theta_\tau^2 \right]^2 \\
    &=  \left[ p_1 + p_2\theta_\tau + p_3 \theta_\tau^2 \right]^2 \\
\end{align}
where we let:
\begin{align}
    p_1 = ab - \alpha (a^2 + b^2)ab + \alpha^2 a^3 b^3,
    \qquad
    p_2 =\alpha (a^2 + b^2) - 2\alpha^2 a^2 b^2,
    \qquad
    p_3 = \alpha^2 ab
\end{align}
We then obtain
\begin{align}
\E f^2(a, b, \theta_\tau)    &= p_1^2 + p_2^2 + 3p_3^2 + 2 p_1 p_3
\end{align}
and where the last equality is obtained be remembering the expectation properties of $\theta_\tau$:
\begin{align}
    \E\nolimits_{p(\tau)} \theta_\tau^2 = 1,
    \qquad
    \E\nolimits_{p(\tau)} \theta_\tau^3 = 0,
    \qquad
    \E\nolimits_{p(\tau)} \theta_\tau^4 = 3.
\end{align}
Furthermore, we obtain
\begin{equation}
\E \theta_\tau f(a, b, \theta_\tau) = p_2
\end{equation}
This gives us a final expression for the MAML loss:
\begin{equation}
    \mathcal{L}^\text{MAML}_{\textsc{deep}} = \frac{1}{2} (1 + p_1^2 + p_2^2 + 3p_3^2 + 2p_1 p_3 - 2p_2).
\end{equation}
We use the Matlab Symbolic Toolbox to help us simplify the rest of the calculation. The code is given below
\lstset{language=Matlab}
{\small
\begin{lstlisting}
%% We use sa, sb, salpha as the "symbolic version" of a, b, and alpha
syms sa sb salpha

pp1 = sa*sb*(1- salpha*(sa^2+sb^2)+ salpha^2*sa^2*sb^2);
pp2 = salpha*(sa^2+sb^2) - 2 *salpha^2 *sa^2*sb^2;
pp3 = salpha^2*sa*sb;

L = pp1*pp1 + pp2*pp2 + 3*pp3*pp3 + 2*pp1*pp3 -2 *pp2;
diffa = diff(L, sa);
diffb = diff(L, sb);

H = [diff(diffa, sa) diff(diffa, sb);diff(diffb, sa) diff(diffb, sb)];
\end{lstlisting}
}
\paragraph{Stationary Points} The gradients are complicated polynomials in $a$ and $b$. Instead, we give the specific cases:
\begin{align}
\left. \parde{\mamlloss_{\textsc{deep}}}{a} \right|_{b=0} = 4 \alpha a (\alpha a^2-1), \quad \left. \parde{\mamlloss_{\textsc{deep}}}{b} \right|_{b=0}   = 0
\end{align}
We immediately can derive that there are at least 5 stationary points:
\begin{itemize}
\item $(a=0, b=0)$, with Hessian 
\begin{equation}
H = \left[\begin{aligned}&-4\alpha & 0 \\ 
& 0  & -4\alpha\end{aligned}\right]
\end{equation}
\item $(a= \pm \frac{1}{\sqrt{\alpha}}, 0)$, with Hessian
\begin{equation}
H = \left[\begin{aligned}&8\alpha & 0 \\ 
& 0  & 6\alpha^3\end{aligned}\right]
\end{equation}
\item $(a= 0, b = \pm \frac{1}{\sqrt{\alpha}})$, with Hessian
\begin{equation}
H = \left[\begin{aligned}&6\alpha^3 & 0 \\ 
& 0  & 8\alpha\end{aligned}\right]
\end{equation}
\end{itemize}

\section{Supplementary Experiments}

This section provides additional experimental evidence to complement the evidence presented in the main text.

\subsection{Binary Logistic Regression}

To resonate more with our experiments of using (multinominal) logistic regression models on Omniglot, CIFAR and MNIST datasets, we also analyze binary logistic regression models on synthetic data.

\begin{figure}[ht]
    \centering
    \includegraphics[height=0.32\textwidth]{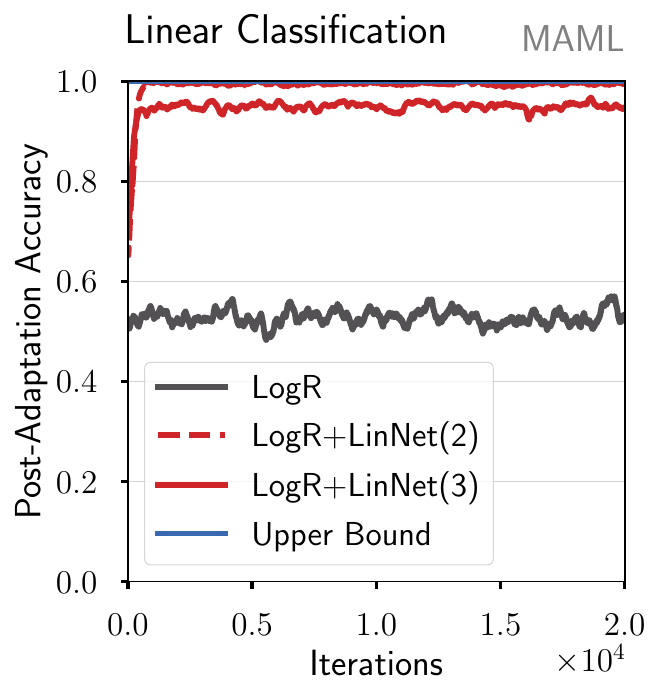}
    \includegraphics[height=0.32\textwidth]{binary_experiments/accuracy_plots/neurips-with-collapse.pdf}
    \caption{\small Meta-learning of linear models on synthetic data. For linear separable data, MAML fails on logistic regression but succeed on logistic regression augmented with 2 or 3 linear layers. \textbf{(Left)} Meta-train accuracies. \textbf{(Right)} Meta-test accuracies.
    } \label{fLogistic}
\end{figure}

\subsubsection{MAML on Logistic Regression and Logistic Regression with Linear Layers}

We randomly sample a set of 2-dimensional task parameters $\theta_\tau  \in \R^{2}$ from a standard multivariate spherical Gaussian and use each of them to define a linear decision boundary of a binary classification task. We sample inputs from a 2-dimensional multivariate spherical Gaussian and the binary outputs for each task are sampled from $y\sim \sigma(\theta_\tau\T\vx)$, where $\sigma()$ is the sigmoid function.

By construction, a logistic regression (LR) is sufficient to achieve very high accuracy on any task.  But can MAML learn a linear classifier from a randomly sampled subset of training tasks that is able to adapt quickly to the test tasks? It is intuitive to see the minimizer for the MAML loss $\mamlloss$ is the origin due to the rotation invariances of both the task parameters and the inputs. The origin thus provides the best initialization to adapt to new tasks  by not favoring any particular task. 

Fig.~\ref{fLogistic}(Left) reports the 1-step post-adaptation accuracy on the test tasks for the meta-learned logistic regression model. Equally surprising as the previous results, the model fails to perform better than chance. However, adding linear layers ``rescues'' the meta-learning; LR with 2 or 3 linear layers attains nearly perfect classification on any task.

\subsection{Logistic Regression Failure Modes}

\begin{figure}[ht]
    \centering
    \small
    \includegraphics[width=0.32\textwidth]{NEURIPS-OM-5W5S-LINEAR/test.pdf}
    \includegraphics[width=0.32\textwidth]{NEURIPS-CFS-5W5S-LINEAR/test.pdf}
    \includegraphics[width=0.32\textwidth]{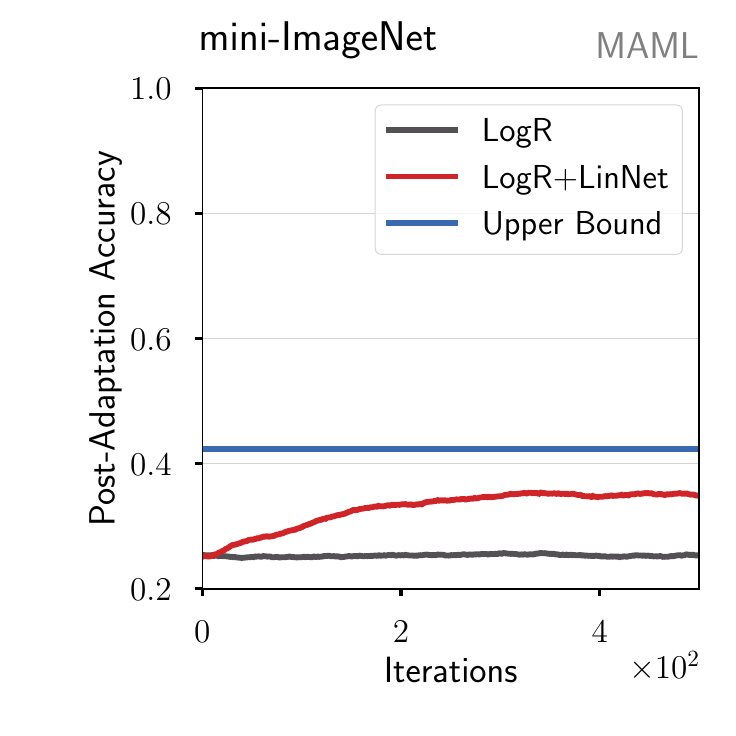} 
    \caption{\small Meta-training logistic regression models with MAML on Omniglot, CIFAR-FS, and mini-ImageNet led to poor performances. Adding linear nets improves meta-learning significantly, \emph{without} changing the model's capacity.}
    \label{fLogRegLinNetSupp}
\end{figure}

As argued in \S5.1, shallow logistic regression models fail to meta-learn on Omniglot and CIFAR-FS.
Fig.~\ref{fLogRegLinNetSupp} displays this same failure mode on the mini-ImageNet dataset.
To simplify computations, we downscaled the original mini-ImageNet images (84x84 pixels, with RGB channels) to the same size as CIFAR-FS. (32x32 pixels, with RGB channels)

\subsection{Linear Layers Cannot Be Collapsed}

Our previous results have shown that when MAML cannot meta-train well, adding multiple linear layers helps to meta-learn, cf. Fig.~\ref{fLogistic} and Fig.~1 of the main text.

\begin{figure}[ht]
    \centering
    \includegraphics[width=0.32\textwidth]{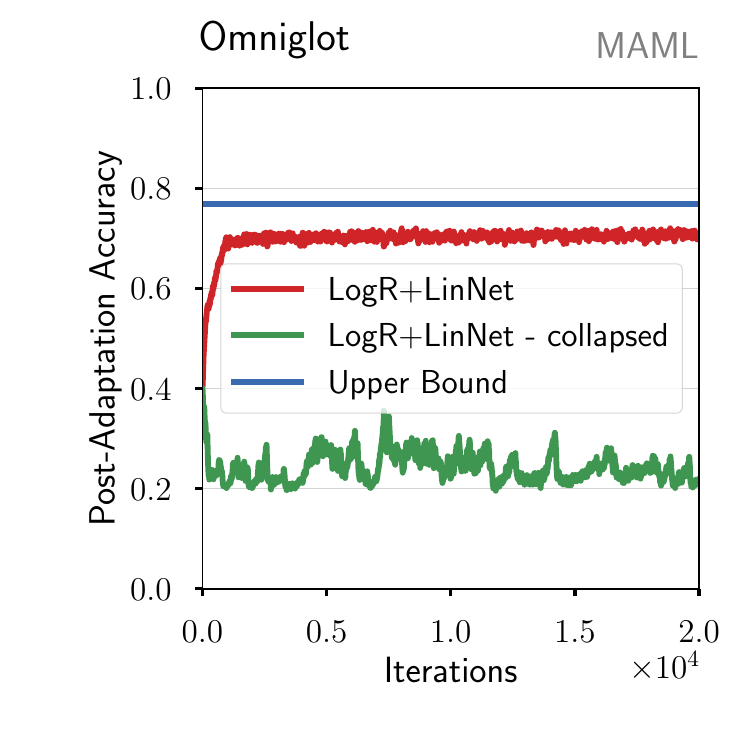}
    \includegraphics[width=0.32\textwidth]{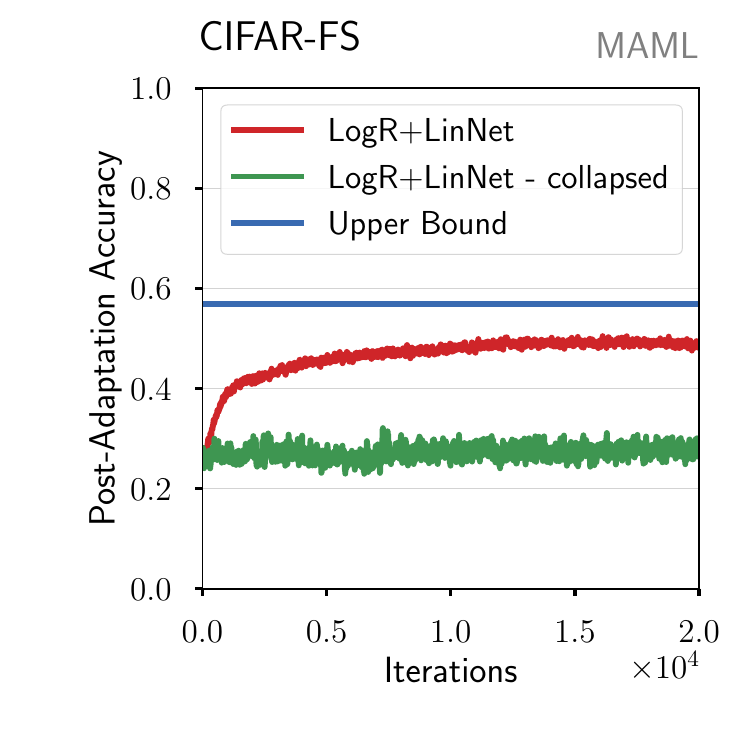}
    \includegraphics[width=0.32\textwidth]{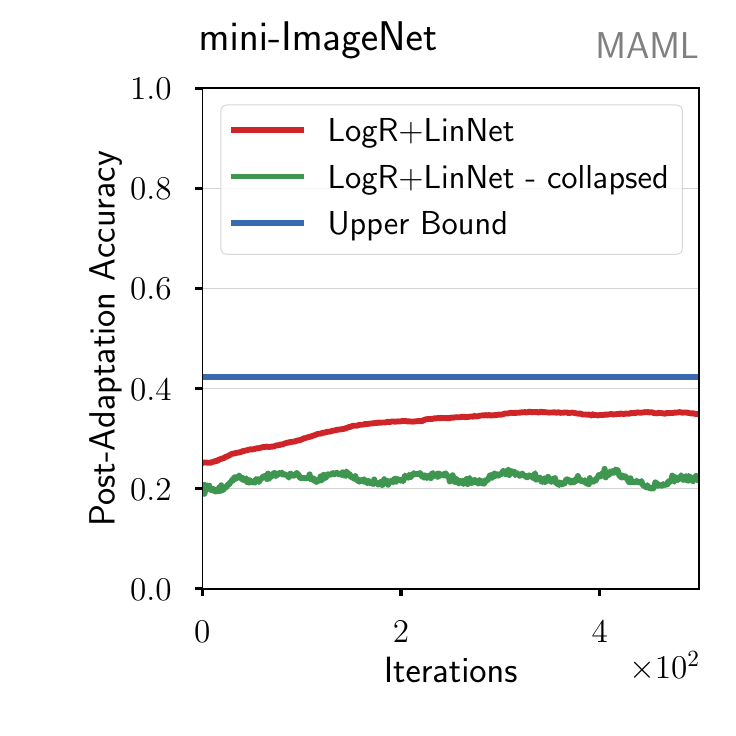}
    \caption{
        Collapsing multiple linear layers into a single layer before adapting to new tasks (green curves) leads to poor adaptation results.
    } \label{fCollapse}
\end{figure}

But after meta-training, can we collapse those multiple linear layers into a single linear layer so as to reduce the model size?

Fig.~\ref{fCollapse} shows that collapsing erases the model's ability to adapt to new tasks.  This suggests that \textbf{the solutions identified in models with additional linear layers need to stay in the overparameterized space to be effective and they cannot be collapsed before adaptation}.

\subsection{Does SGD-induced implicit regularization for linear networks matter for meta-learning?}

Implicit regularization refers to the bias towards solutions with different generalization abilities when stochastic gradient descent (SGD) is used to optimize overparameterized models~\cite{Gidel2019-rk}.

One might want to argue that implicit regularization explains why additional linear layers in our experiments (Fig.~1 of the main text, Fig.~\ref{fLogistic}) help to meta-learn. This is an interesting hypothesis, though we do not think it can fully explain what is observed.

First, as those experiments have shown, for models augmented with linear nets, there is very little difference between training accuracies and meta-testing accuracies. Moreover, for models without linear nets, there is very little difference between the two phases either. Thus, while SGD could lead to solutions with different generalization abilities in regular supervised learning settings, our experiments do not show there is an overfitting issue in the experiments here for meta-learning.

The difference between parsimonious and overparameterized models (with linear layers) could be due to the difference in solutions. However, the solutions are in space of different dimensionality. Fig.~\ref{fCollapse} shows that \textbf{after} we collapse the overparameterize models such that the solutions are in the same space, the solutions from the collapsing becomes as ineffective (in terms of leading to poor adaptation results) as the original models.   This suggests that SGD is \textbf{not} identifying a solution in the overparameterized space that could have been identified in the original space. In other words, SGD is not biasing towards any solution in the overparameterized space that could have made difference in the original space.

\subsection{ANIL with Meta-Optimizers}

In \S5.3, we showed that MAML augmented with meta-optimizers was able to meta-learn in shallow non-linear networks.
Table~\ref{tShallowAdaptationMetaOptimizaerANIL} shows similar results when using ANIL as the meta-learning algorithm.
(Note: this table omits T-Nets, as it is not compatible with ANIL.)
We observe that ANIL always benefits from the combination with a meta-optimizer, and specifically that it benefits most from KFO.

\begin{table}[ht]
\vspace{0em}
\centering
\caption{Meta-Optimizers Improve ANIL Meta-Learnability on CNN(2)}
\label{tShallowAdaptationMetaOptimizaerANIL}
\setlength{\tabcolsep}{3pt}
{\small
\begin{tabular}{l ccccc}
\toprule
Dataset          & ANIL &  \multicolumn{3}{c}{ANIL w/} \\
\cmidrule{3-5}
              &       & MSGD  & MC    & KFO \\
\midrule
Omniglot      & 91.00 & 92.47 & 92.67 & \textbf{94.40}\\
CIFAR-FS      & 66.10 & 67.55 & 67.43 & \textbf{68.81} \\
mini-ImageNet & 56.42 & 57.45 & 59.07 & \textbf{62.65}\\

\bottomrule
\end{tabular}
}
\vspace{-0em}
\end{table}

\subsection{Effect of Number of Layers}

\begin{figure}[ht]
    \begin{center}
        \includegraphics[width=0.32\textwidth]{layers_vs_acc_5w5s/omniglot.pdf}      
        \includegraphics[width=0.32\textwidth]{layers_vs_acc_5w5s/cifarfs.pdf}
        \includegraphics[width=0.32\textwidth]{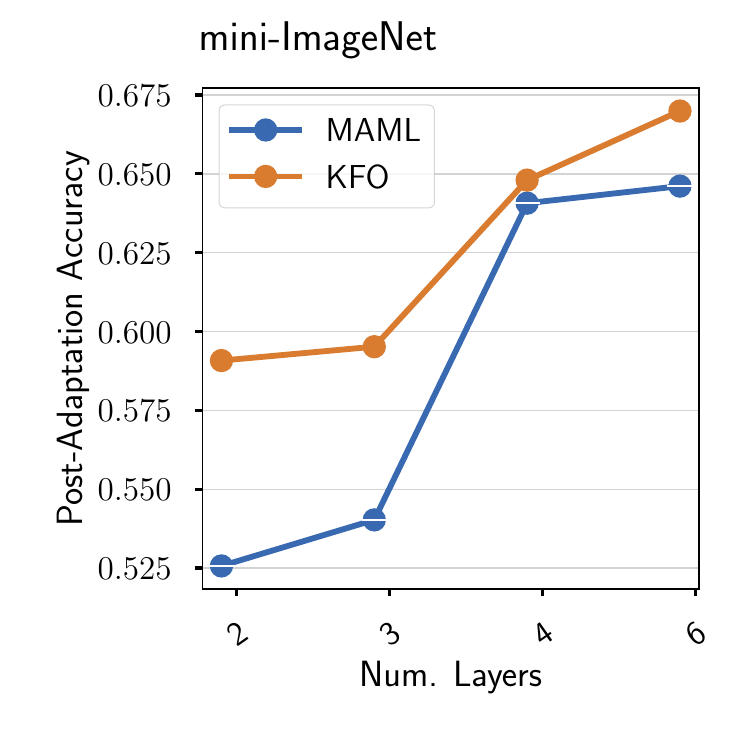}
    \end{center}
    \caption{\small 
    Complementing Fig.~\ref{fig:exp-layers}, we vary the number of convolutional layers in the model.
    As the model size increases the performance of both methods improves, which can be attributed to the model's increased capacity to learn the target task.
    However, the net gain from \textsc{meta-kfo} over MAML has diminishing returns as the number of layers increases since the benefit of the external meta-optimizer reduces as the upper layers of the larger models have more capacity to meta-learn to control their own bottom layers.
    }
    \label{fig:supp-exp-layers}
\end{figure}

Figure~\ref{fig:supp-exp-layers} complements the results in \S5.3 when varying the number of layers in the convolutional network.
(We include again results on Omniglot and CIFAR-FS for convenience.)
As we observed in the main text, shallower models greatly benefit from the additional meta-optimization parameters which alleviates the burden of learning how to adapt.
But, this benefit dampens as we increase the number of convolutional layers -- the gap between \textsc{meta-kfo} and MAML shrinks -- since the additional upper layers implicitly transform the gradient of lower layers, thus enabling successful meta-learning.

\section{Details on KFO}

\subsection{Other factorization schemes}

A technical issue arises when expressing the meta-optimizer $U_\xi$ as a neural network: the dimensionality of modern (model) network architectures ranges in the tens of thousands, if not more.
To address those computational and memory issues, we learn one meta-optimizer network per matrix parameter in the model network and express the weights of the optimizer neural network as a Kronecker factorization such that for each weight $W \in \R^{m \times n}: W = R^\top \otimes L$, where $R \in \R^{m \times m}$ and $L \in \R^{n \times n}$.

Many matrix factorization schemes could be used  and would result in different modeling and computational trade-offs.
For example, using a low-rank Cholesky factorization $A = LL^T$ where $L \in \R^{k \times r}$ allows to interpolate between computational complexity and decomposition rank by tuning the additional hyper-parameter $r$.
The Cholesky decomposition might be preferable to the Kronecker one in memory-constrained applications, since $r$ can be used to control the memory requirements of the meta-optimizer.
Moreover, such a decomposition imposes symmetry and positiveness on $A$, which might be desirable when approximating the Hessian or its inverse.

In this work, we preferred the Kronecker decomposition over alternatives for three reasons: (1) the computational and memory cost of the Kronecker-product are acceptable,
(2) $R^\top \otimes L$ is full-rank whenever $L, R$ are full-rank, and (3) the identity matrix lies in the span of Kronecker-factored matrices.
In particular, this last motivation allows meta-optimizers to recover the gradient descent update by letting $R, L$ be the identity.

\subsubsection{Schematics and Pseudo-code}
\begin{figure}
\begin{center}
    \includegraphics[width=0.75\linewidth]{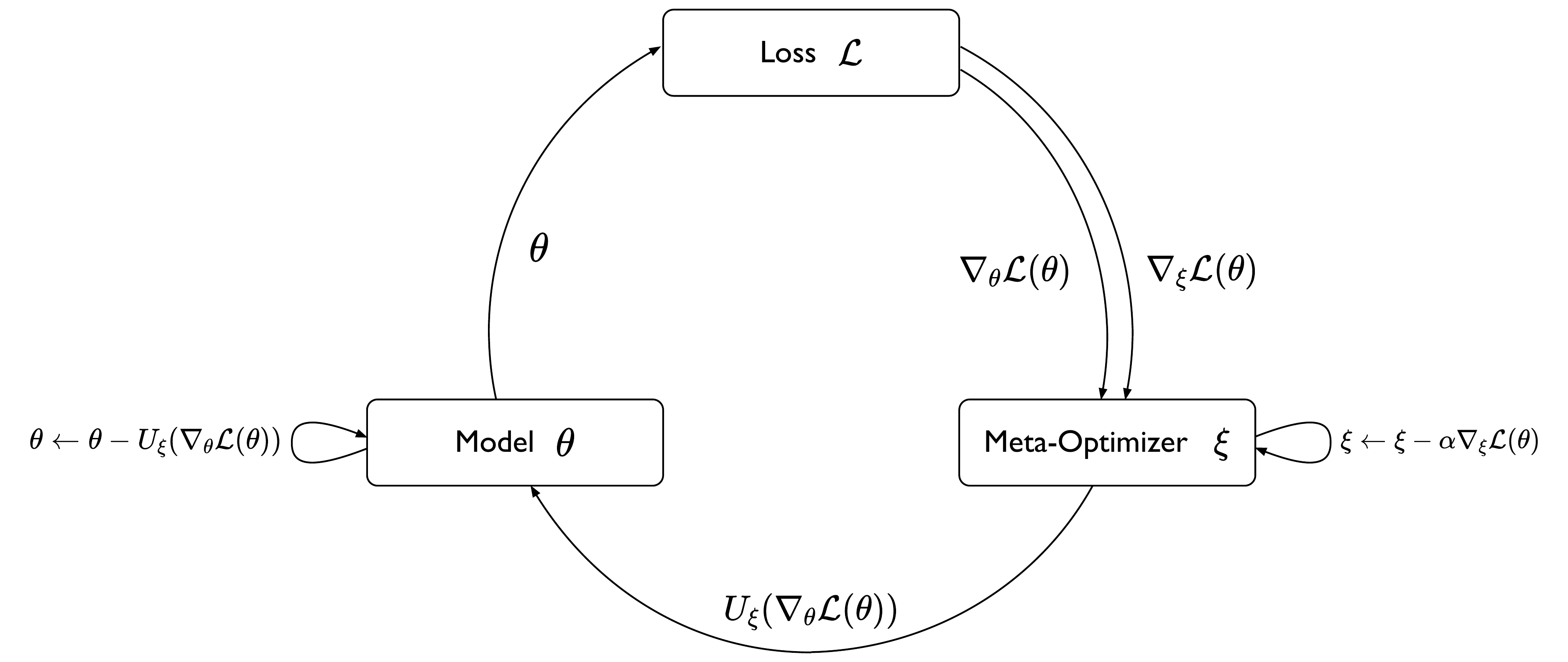}
\end{center}
\caption{Schematic of model-optimizer loop}
\label{fModelOptimLoop}
\end{figure}

Pseudo-code for meta-optimizers is provided in Algorithm~\ref{aMetaOpt}, and a schematic of the model-optimizer loop in Figure~\ref{fModelOptimLoop}.

\begin{algorithm*}[h] 
    \caption{Meta-Learning with Meta-Optimizers}
    \label{aMetaOpt}
    \begin{algorithmic}[1] 
        \Require Fast learning rate $\alpha$, Initial parameters $\theta^{\text{(init)}}$ and $\xi^{\text{(init)}}$, Optimizer \texttt{Opt}
        \While{$\theta^\text{(init)}, \xi^\text{(init)}$ not converged}
        \State Sample task $\tau \sim p(\tau)$
        \State $\theta_1 = \theta^\text{(init)}, \quad \xi_1 = \xi^\text{(init)}$
        \For{step $t = 1, \dots, T$}
        \State Compute loss $\loss_\tau(\theta_t)$
        \State Compute gradients $\nabla_{\theta_t} \loss_\tau(\theta_t)$ and $\nabla_{\xi_t} \loss_\tau(\theta_t)$
        \State Update the meta-optimizer parameters $\xi_{t+1} = \xi_t - \alpha \nabla_{\xi_t} \loss_\tau(\theta_t)$
        \State Compute the model update $U_{\xi_{t+1}}(\nabla_{\theta_t} \loss_\tau(\theta_t))$
        \State Update the model parameters $\theta_{t+1} = \theta_t - U_{\xi_{t+1}}(\nabla_{\theta_t} \loss_\tau(\theta_t))$
        \EndFor
        \State Update model and meta-optimizer initializations
        \State \qquad $\theta^\text{(init)} \gets \theta^\text{(init)} - \texttt{Opt}(\nabla_{\theta^\text{(init)}} \loss_\tau(\theta_T))$
        \State \qquad $\xi^\text{(init)} \gets \xi^\text{(init)} - \texttt{Opt}(\nabla_{\xi^\text{(init)}} \loss_\tau(\theta_T))$
        \EndWhile
    \end{algorithmic}
\end{algorithm*}

\section{Hyperparameters and details for reproducibility}

We report the hyper-parameter values for the experiments of the main text.
Each experimental setup is tuned independently based on post-adaptation accuracies obtained on validation tasks.

All experiments are implemented on top of PyTorch~\cite{Paszke2019-fx} and learn2learn~\cite{Arnold2020-ss}.
Moreover, our work relied on tools from the Python scientific ecosystem~\cite{Van_Rossum1995-za}, including numpy~\cite{Hunter2007-ji}, matplotlib~\cite{Harris2020-uq}, and scipy~\cite{Virtanen2020-dv}.
We provide example implementations of Meta-KFO at: \url{https://github.com/Sha-Lab/kfo}

\paragraph{Linear Models}
On the 1D linear regression, both \textsc{shallow} and \textsc{deep} models use a fast-adaptation learning rate set to 0.1.
Their parameters are optimized by SGD with a meta learning rate set to 0.01, and a momentum term set to 0.9.
On the vision datasets (Omniglot, CIFAR-FS, mini-ImageNet), the meta and adaptation learning rates are given in Table~\ref{tLinearLearningRates}.
For those experiments, the added linear network consists of 256-128-64-64 hidden units.
To simplify computations on mini-ImageNet, we downsample images to 32x32 pixels.

\begin{table}[t]
    \centering
    \caption{\textbf{Adaptation learning rates} for the linear model experiments in the main text and \supp }
    \label{tLinearLearningRates}
    \setlength{\tabcolsep}{3pt}
    {\small
    \begin{tabular}{l l c c}
        \addlinespace

        \toprule
        Dataset       & Model        & Meta learning rate & Adaptation learning rate \\
        \midrule
        Synthetic     & LR           & 0.01               & 0.900 \\
        Synthetic     & LR+LinNet(2) & 0.1                & 0.900 \\
        Synthetic     & LR+LinNet(3) & 0.1                & 0.900 \\
        \midrule
        Omniglot      & LR           & 0.0005             & 1.0 \\
        Omniglot      & LR+LinNet    & 0.0005             & 0.08 \\
        \midrule
        CIFAR-FS      & LR           & 0.0005             & 0.01 \\
        CIFAR-FS      & LR+LinNet    & 0.0001             & 0.02 \\
        \midrule
        mini-ImageNet & LR           & 0.0005             & 0.01 \\
        mini-ImageNet & LR+LinNet    & 0.0005             & 0.01 \\
        \bottomrule

    \end{tabular}
    }
\end{table}

\paragraph{Nonlinear Models}
All methods in Tables~1, ~2, and~3 are tested on the standard 5-ways 5-shots setting.
For Omniglot and CIFAR-FS, the networks are the original CNN from \cite{Finn2017-an} with 32 hidden units, while for mini-ImageNet we used 64 hidden units.
We use Adam to optimize the networks, with default values of (0.9, 0.999) for $\beta$ and 1e-8 for $\epsilon$, and process tasks by batches of 16.
Meta and adaptation learning rates for all methods are reported in Tables~\ref{tShallowAdaptationAdapt},~\ref{tShallowAdaptationMeta}.

\begin{table}[h]
    \centering
    \caption{\textbf{Adaptation learning rates} for the two-layer CNN experiments in the main text and \supp}
    \label{tShallowAdaptationAdapt}
    \setlength{\tabcolsep}{3pt}
    {\small
    \begin{tabular}{l ccccc c  cccccc}
        \addlinespace
        \toprule
        & \multicolumn{5}{c}{MAML w/ } &        & \multicolumn{5}{c}{ANIL w/ }\\
        \cmidrule{2-6} \cmidrule{8-12}

        Dataset       & (null) & LinNet & MSGD & MC   & KFO   &  & (null) & LinNet & MSGD & MC   & KFO \\
        \midrule
        Omniglot      & 0.08   & 0.05   & 0.5  & 0.5  & 0.1   &  & 0.5    & 0.5    & 0.05 & 0.05 & 0.1\\
        CIFAR-FS      & 0.07   & 0.01   & 0.7  & 0.7  & 0.1   &  & 0.5    & 0.5    & 0.1  & 0.1  & 0.1\\
        mini-ImageNet & 0.001  & 0.001  & 0.1  & 0.05 & 0.001 &  & 0.5    & 0.5    & 0.1  & 0.1  & 0.1\\

        \bottomrule
    \end{tabular}
    }
\end{table}

\begin{table}[h]
    \centering
    \caption{\textbf{Meta learning rates} for the two-layer CNN experiments in the main text and \supp }
    \label{tShallowAdaptationMeta}
    \setlength{\tabcolsep}{3pt}
    {\small
    \begin{tabular}{l ccccc c  cccccc}
        \addlinespace
        \toprule
        & \multicolumn{5}{c}{MAML w/ } &        & \multicolumn{5}{c}{ANIL w/ }\\
        \cmidrule{2-6} \cmidrule{8-12}

        Dataset       & (null) & LinNet & MSGD   & MC     & KFO    &  & (null) & LinNet & MSGD  & MC    & KFO \\
        \midrule
        Omniglot      & 0.003  & 0.003  & 0.0005 & 0.0005 & 0.001  &  & 0.001  & 0.003  & 0.001 & 0.001 & 0.001\\
        CIFAR-FS      & 0.003  & 0.003  & 0.001  & 0.003  & 0.003  &  & 0.002  & 0.001  & 0.001 & 0.001 & 0.001\\
        mini-ImageNet & 0.0005 & 0.0005 & 0.001  & 0.001  & 0.0001 &  & 0.001  & 0.001  & 0.001 & 0.001 & 0.001\\

        \bottomrule
    \end{tabular}
    }
\end{table}

\clearpage

\end{document}